# Direct Processing of Document Images in Compressed Domain

Mohammed Javed*[1], P. Nagabhushan*[2], B.B. Chaudhuri[#3]

*Department of Studies in Computer Science, University of Mysore, Mysore – 570006, India
#Computer Vision and Pattern Recognition Unit, Indian Statistical Institute, Kolkata-700108, India
Email: [1]javedsolutions@gmail.com, [2]pnagabhushan@hotmail.com, [3]bbc@isical.ac.in

**Abstract:** With the rapid increase in the volume of Big data of this digital era, fax documents, invoices, receipts, etc are traditionally subjected to compression for the efficiency of data storage and transfer. However, in order to process these documents, they need to undergo the stage of decompression which indents additional computing resources. This limitation induces the motivation to research on the possibility of directly processing of compressed images. In this research paper, we summarize the research work carried out to perform different operations straight from run-length compressed documents without going through the stage of decompression. The different operations demonstrated are feature extraction; text-line, word and character segmentation; document block segmentation; and font size detection, all carried out in the compressed version of the document. Feature extraction methods demonstrate how to extract the conventionally defined features such as projection profile, run-histogram and entropy, directly from the compressed document data. Document segmentation involves the extraction of compressed segments of text-lines, words and characters using the vertical and horizontal projection profile features. Further an attempt is made to segment randomly a block of interest from the compressed document and subsequently facilitate absolute and relative characterization of the segmented block which finds real time applications in automatic processing of Bank Cheques, Challans, etc, in compressed domain. Finally an application to detect font size at text line level is also investigated. All the proposed algorithms are validated experimentally with sufficient data set of compressed documents.

## 1. Introduction

Rapid growth of multimedia data such as fax documents, invoices, receipts, images, audios and videos over the internet, digital libraries and e-governance applications have traditionally resulted in compressing the data for the efficiency of data storage and transfer. However, in order to process them or carry out any analytical computations, they need to undergo the stage of decompression, which indents additional computing resources. This limitation incites the motivation to research on the possibility of directly processing of compressed data. This alternate, challenging and breakthrough idea of operating directly over the compressed representation without using decompression, is called *compressed domain processing*.

Direct analysis of compressed data could open up a newer vista in digital image analysis. In this direction, the proposed research work aims at exploring the possibility of operating over specifically the compressed document images available in large databases and many network applications. In the sequel, we also aim to assess the advantages and limitations of such direct processing without involving full or total decompression.

Our research study aims at the exploration for basic operations like segmentation, extraction of objects of interest and their related analytics straight from compressed documents. For all these operations, we probe for a suitable feature extraction mechanism with the available techniques such as profiling, histogram and entropy. For the purpose of validation and corroboration of our proposed methods, we propose to experiment on large sets of compressed documents.

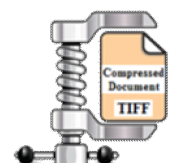

Rest of the paper is organized as follows: section-2 describes a detailed survey on document processing in compressed domain, section-3 gives brief idea about the contributions[14-18] made by us so far in the area of proposed research, and section-4 summarizes the entire work and gives directions for further research in the area.

## 2. Document Processing in Compressed Domain

Although image, audio and video data are rendered in compressed form for the efficiency of data storage and transfers, the existing methods still need to decompress and operate on the data. Since, the data exists in compressed form; it shall be a novel to think of performing operations directly on the compressed data, bypassing the stage of total decompression. Therefore, this research is an attempt to explore this novel idea, taking compressed documents to perform the existing operations directly without decompressing the data.

There are many document image compression file formats available in the literature: BMP, PDF, JPEG and TIFF are commonly preferred formats. TIFF is very popular in handling fax and handwritten documents and also preferred for archiving in digital libraries and in many network applications like printers, fax and xerox machines.

TIFF provides many built in compression algorithms: CCITT Group 3 (T.4), CCITT Group 4 (T.6) and many more. T.6 provides better compression ratio than T.4 and hence it was developed for archival purpose. However, the encoding process in T.6 is 2D and takes previous encoded line to encode the current line. In case of any single error during transmission, the whole document becomes unusable after decoding. Therefore, the natural choice for fax or network applications is T.4 standard, which facilitates both 1D and 2D coding. In this research work, we propose to use data from 1D (line by line) coding popularly known as Modified Huffman (MH) coding which is supported by all fax compression techniques. The basic and lossless compression scheme used here is Run-Length Encoding (RLE). As we focus our study on text documents, we limit our work to TIFF compressed, run-length encoded binary documents.

The earliest idea of operating directly on run-length encoded compressed documents can be related to the work of [1] and [2]. RLE, a simple compression method was first used for coding pictures [3] and television signals [4]. There are several efforts in the direction of directly operating on document images using run-length code. Operations like image rotation [5], connected component extraction [6], skew detection [7], page layout analysis [8], bar code detection [9] are reported in the literature. There are also some initiatives in finding document similarity [10], equivalence [11] and retrieval [12]. One of the recent works using run-length information is to perform morphological related operations [13]. In most of the works, they use either run-length information from the uncompressed image or do some decoding to perform operations. However, to our best knowledge a detailed study on compressed documents from the viewpoint of computational cost, efficiency and validation with large dataset has not been attempted in the literature. On the other hand, research on compressed documents has been an area of interest to both industry and academic groups. However, there has been no much significant progress in this direction, perhaps due to the complex nature of compressed data. Therefore, in this backdrop the proposed research study is dedicated specifically to explore the possibility of implementing some of the aforementioned operations on document images in compressed domain.

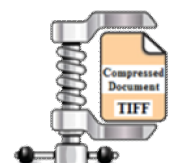

## 3. Direct Processing of Run-length Compressed Document Data

The basic compression technique behind TIFF compressed text and fax documents is Run Length Encoding (RLE). In RLE, a run is a sequence of pixels having similar value and the number of such pixels is length of the run. The representation of the image pixels in the form of sequence of run values ($P_i$) and its lengths ($L_i$) is run-length encoding. It can be mathematically represented as ($P_i, L_i$). However, for binary images the coding is done with alternate length information of black and white pixels, which results in compact representation. The Table-1, gives the description of compressed data using RLE technique. The compressed data consists of alternate columns of number of runs of '0' and '1' identified as odd columns (1,3,5,…) and even columns (2,4,6…) respectively. The proposed algorithms in this paper, work on these types of data that will be frequently referred in this paper as compressed data.

TABLE I: Description of run-length compressed binary data

| Line | Binary image Data | 1 | 2 | 3 | 4 | 5 |
|---|---|---|---|---|---|---|
| 1: | 00000000000000 | 14 | 0 | 0 | 0 | 0 |
| 2: | 00110000111110 | 2 | 2 | 4 | 5 | 1 |
| 3: | 01111000111110 | 1 | 4 | 3 | 5 | 1 |
| 4: | 01111000111110 | 1 | 4 | 3 | 5 | 1 |
| 5: | 01111000111110 | 1 | 4 | 3 | 5 | 1 |
| 6: | 00110000000000 | 2 | 2 | 10 | 0 | 0 |
| 7: | 10000000000000 | 0 | 1 | 13 | 0 | 0 |
| 8: | 10000000000000 | 0 | 1 | 13 | 0 | 0 |
| 9: | 00100001111100 | 2 | 1 | 4 | 5 | 2 |
| 10: | 01110001111100 | 1 | 3 | 3 | 5 | 2 |
| 11: | 01111001111100 | 1 | 4 | 2 | 5 | 2 |
| 12: | 01111100000000 | 1 | 5 | 8 | 0 | 0 |
| 13: | 00000000000000 | 14 | 0 | 0 | 0 | 0 |

**3.1 Feature extraction: Projection profile, run-histogram and entropy**

In this section, we present the basic ideas of extracting the existing conventional features straight from the compressed data and subsequently using it for compressed document segmentation. More details regarding the methods and mathematical formulations can be obtained from our work in [14] [15] and [16].

**3.1.1 Projection Profile**

A projection profile is a histogram of the number of black pixel values accumulated along parallel lines taken through the document. In the compressed data, the presence of alternate runs of black and white pixels hints us to add row-wise the alternate black-pixel-runs and skipping all white-pixel-runs to obtain Vertical Projection Profile (VPP). The proposed method is simple and straight forward in getting VPP curve directly from the compressed data. However, obtaining Horizontal Projection Profile (HPP) curve is a complex task. This is because the vertical information is not directly available in the compressed representation. So, an algorithm is developed to compute HPP by pulling-out the run values from all the rows simultaneously using first two columns from compressed data as shown in Table-1. In presence of zero run values in both the columns, the runs on the right are shifted to two positions leftwards. Thus for every pop operation, the run value is decremented by 1 and if the popped-out element is from first column then the transition value is 0 otherwise 1. This process is repeated for all the rows generating a sequence of column transitions from the compressed file that may be called virtual decompression. Thus addition of all these popped-out elements column-wise results in HPP curve.

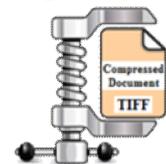

### 3.1.2 Run-Histogram

Generally a histogram of a digital image represents the frequency of number of intensity values present in the image. However for binary images, a simple histogram would not give much information needed for document analysis. This motivates us to define run-histogram on compressed data, which has been recently applied for document retrieval and classification [19] using uncompressed documents. The compressed data in Table-1 contains alternate columns of white and black pixel runs. Using this data structure, we compute the frequency of each run separately for even and odd columns to obtain black-pixel and white-pixel run-histograms. Combining these two run-histograms we get black-white run-histogram. Very recently the work of [19] incited us to extend the idea of run-histogram to logarithmic scale run-histogram. The quantized length of the runs in a logarithmic scale is as follows: [1],[2],[3-4],[5-8],[9-16],…,[129-]. Further details are available in [14].

### 3.1.3 Entropy

Here, we demonstrate the extraction of Conventional Entropy Quantifier (CEQ) and Sequential Entropy Quantifier (SEQ) proposed by [21], straight from the compressed data. The details regarding the idea, motivation and formulation of CEQ and SEQ can be obtained in [21]. CEQ measures the energy contribution of each row by considering the probable occurrence of +ve (0-1) and -ve (1-0) transitions among the total number of pixels present in that particular row. On the other hand SEQ measures the entropy at the position of occurrence of these transitions. However from the compressed data, these information are easily available in the alternate even (+ve) and odd (-ve) columns as shown in Table-1. In each row, counting the number of even or odd columns of non-zero runs, leaving the first column (run) gives the number of +ve or -ve transitions respectively. Whereas, the summation of all the previous runs incremented by 1 is the position of the transition point at a particular column as shown in Table-1. For more details refer [14][15].

### 3.2 Entropy computations and analysis

In this work, we extend the computation of proposed entropy features CEQ and SEQ for compressed documents in both horizontal and vertical directions. Because, the run-length data is horizontally compressed, the vertical information is not directly available for vertical entropy computation. Therefore, the vertical information is generated by popping-out the runs in an intelligent sequence and vertical entropy is computed as detailed in [15]. The entropy computation is demonstrated over the text-line, word and character levels to show the effectiveness of the computed features and their possible application in document image analysis. More details regarding the experimentation and analysis is available in [15].

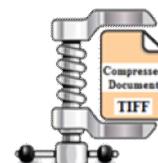

**3.3 Text-line, word and character segmentation**

We present here, the applications of proposed projection profile features (vertical and horizontal) from compressed data to segment a text document into text-lines, words and characters [16].

The compressed data in Table-1 shows rows of alternate columns of runs of white and black pixels. In absence of any black pixel in a row, the compressed row will contain only one run of white pixels whose size will be equal to the width the original document. This special feature of compressed data can be seen in line numbers 1 and 13 of Table-1, which enables us to use projection profile technique for line segmentation in printed text-documents. With our observation, in order to do line segmentation using compressed data, obtaining VPP is sufficient. Making use of minima points in the VPP curve, we trace the vertical start and end points of the text lines. The difference between the end point of previous line and start point of current line is difference between the text lines.

For printed documents, the common approach used for word and character segmentation is HPP. For compressed data, HPP requires the usage of virtual decompression algorithm which is already discussed above. A segmented text-line is an input to this method. The transition values popped out from each row for every column of the compressed text-line are used to trace the characters from left to right. This virtual line with zero transitions hints for the presence of space between the characters. Using this space information we perform character and word segmentation. Once a vertical space between the characters is detected, the space-counter starts incrementing until a non-space column is detected. If the space-counter is less than the defined word-space-threshold, then it is identified as a character and entry is made into the character segmentation matrix. In case the condition is false then a word is detected which is stored in word segmentation matrix. This process is repeated until all the columns in a segmented line get exhausted. Nevertheless, detailed research work regarding word and character segmentation with a schematic block diagram is available in [16].

**3.4 Detection of font size at text-line level**

Automatic detection of font size finds many applications in the area of intelligent OCRing and document image analysis. Therefore, we present a novel idea of learning and detecting font size directly from run-length compressed text documents at line level using simple line height features, which paves the way for intelligent OCRing and document analysis directly from compressed documents. In the proposed model, the given mixed-case text documents of different font size are segmented into compressed text lines and the features extracted such as line height and ascender height are used to capture the pattern of font size in the form of a regression line, using which the automatic detection of font size is done during the recognition stage.

The method is experimented with a dataset of 50 compressed documents consisting of 780 text lines of single font size and 375 text lines of mixed font size resulting in an overall accuracy of 99.67%. More details regarding the work and experimental results is available in [18]

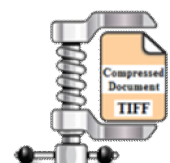

### 3.5 Segmentation and characterization of a text block

This research work introduces a novel method for extracting the specified document block in rectangular segments directly from the run-length compressed document data without going through the stage of decompression. Apart from facilitating extraction of a specified document block, the work is extended further to provide absolute and relative characterization of the extracted blocks using density and entropy features. This research study also demonstrates that document analysis is possible at sub-document level within the compressed document data without decompression and also opens up the gateway for plenty of research issues in the proposed compressed domain. A detailed study on segmentation of a randomly selected text-block and its characterization is available in [17].

### 4. Conclusion and future work plan

In this research work, a novel idea of processing document images directly in their respective compressed domain using the run-length compressed data is demonstrated. Different operation such feature extraction, segmentation, font size detection and characterization of the compressed document is performed directly in run-length compressed data of the document which avoid decompression and results in saving of considerable amount of computing resources. The proposed methods are validated experimentally with sufficient dataset of compressed documents. Finally, this research is a small attempt to demonstrate direct operations with compressed data; however, many open research issues still exists such as handwritten document segmentation, word spotting, numeral spotting, symbol spotting, document retrieval etc. which could be taken as an extension work to this research.

### References

[1] G. Grant and A. Reid, "An efficient algorithm for boundary tracing and feature extraction," Computer Graphics and Image Processing, vol. 17, pp. 225–237, November 1981.

[2] T. Tsuiki, T. Aoki, and S. Kino, "Image processing based on a runlength coding and its application to an intelligent facsimile," Proc. Conf. Record, GLOBECOM '82, pp. B6.5.1–B6.5.7, November 1982.

[3] J. Capon, "A probabilistic model for run-length coding of pictures," IRE Transactions on Information Theory, vol. 5, pp. 157–163, 1959.

[4] J. Limb and I. Sutherland, "Run-length coding of television signals," Proceedings of IEEE, vol. 53, pp. 169–170, 1965.

[5] Y. Shima, S. Kashioka, and J. Higashino, "A high-speed rotation method for binary images based on coordinate operation of run data," Systems and Computers in Japan, vol. 20, no. 6, pp. 91–102, 1989.

[6] C. Ronse and P. Devijver, Connected Components in binary images: The Detection Problem. Research Studies Press,1984.

[7] J. Kanai and A. D. Bangdanov, "Projection profile based skew estimation algorithm for jbig compressed images," International Journal on Document Analysis and Recognition (IJDAR'98), vol. 1, pp. 43–51, 1998.

[8] T. Pavlidis, "A vectorizer and feature extractor for document recognition," Computer Vision, Graphics, and Image Processing, vol. 35, pp. 111–127, 1986.

[9] C. Maa, "Identifying the existence of bar codes in compressed images," CVGIP: Graphical Models and Image Processing, vol. 56, p. 352356, July 1994.

[10] J. J. Hull, "Document matching on ccitt group 4 compressed images," SPIE Conference on Document Recognition IV, pp. 8–14, Feb 1997.

[11] J. J. Hull, "Document image similarity and equivalence det," International Journal on Document Analysis and Recognition (IJDAR'98), vol. 1, pp. 37–42, 1998.

[12] Y. Lu and C. L. Tan, "Document retrieval from compressed images," Pattern Recognition, vol. 36, pp. 987–996, 2003.

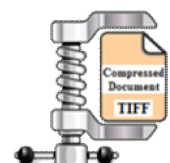

[13] T. M. Breuel, "Binary morphology and related operations on run-length representations," International Conference on Computer Vision Theory and Applications - VISAPP, pp. 159–166, 2008.

[14] Mohammed Javed, P. Nagabhushan, and B.B. Chaudhuri, "Extraction of projection profile, run-histogram and entropy features straight from run-length compressed Documents," Second IAPR Asian Conference on Pattern Recognition (ACPR'13), Nov 5-8, 2013, Okinawa, Japan

[15] P. Nagabhushan, Mohammed Javed, B.B.Chaudhuri, " Entropy Computation of Document Images in Run-length Compressed Domain", (ICSIP'14, Jan 8-10, 2014, Bangalore, India

[16] Mohammed Javed, P. Nagabhushan, and B.B. Chaudhuri, "Extraction of line-word-character segments directly from run-length compressed printed text-documents", NCVPRIPG'13, December 19-21, 2013.

[17] Mohammed Javed, P. Nagabhushan, and B.B. Chaudhuri ,Direct Processing of Run-Length Compressed Document Image for Segmentation and Characterization of a Specified Block, IJCA, vol-83(15), Pages 1-8, December 2013, Published by Foundation of Computer Science, New York, USA.

[18] Mohammed Javed, P. Nagabhushan, and B.B. Chaudhuri, Automatic Detection of Font Size Straight from Run Length Compressed Text Documents, IJCSIT, vol-5(1), Pages 818-825, February 2014, Tech Science Publications.

[19] A. Gordo, F. Perronnin, and E. Valveny, "Large-scale document image retrieval and classification with runlength histograms and binary embeddings," Pattern Recognition, vol. 46, pp. 1898–1905, July 2013.

[20] S. D. Gowda and P. Nagabhushan, "Establishing equivalence between two layout-variant documents images without reading," International Conference on Modeling and Simulation, Kolkata, India, vol. 1, pp. 156–161, 2007.

[21] S. D. Gowda and P. Nagabhushan, "Entropy quantifiers useful for establishing equivalence between text document images," ICCIMA, pp. 420 – 425, 2007.

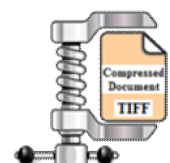